\documentclass{article}

\usepackage[preprint]{neurips_2023}

\usepackage[utf8]{inputenc} 
\usepackage[T1]{fontenc}    
\usepackage{hyperref}       
\usepackage{url}            
\usepackage{booktabs}       
\usepackage{amsfonts}       
\usepackage{nicefrac}       
\usepackage{microtype}      
\usepackage{xcolor}         
\usepackage{times}
\usepackage{tikz}
\usepackage{multicol}
\usepackage{geometry}
\usepackage{changepage}

\title{Turning large language models into cognitive models}

\author{%
  Marcel Binz \\
  MPRG Computational Principles of Intelligence\\
  Max Planck Institute for Biological Cybernetics, T\"ubingen, Germany\\
  \texttt{marcel.binz@tue.mpg.de} \\
   \And
    Eric Schulz \\
  MPRG Computational Principles of Intelligence\\
  Max Planck Institute for Biological Cybernetics, T\"ubingen, Germany
}

\begin{document}

\maketitle

\begin{abstract}
  Large language models are powerful systems that excel at many tasks, ranging from translation to mathematical reasoning. Yet, at the same time, these models often show unhuman-like characteristics. In the present paper, we address this gap and ask whether large language models can be turned into cognitive models. We find that -- after finetuning them on data from psychological experiments -- these models offer accurate representations of human behavior, even outperforming traditional cognitive models in two decision-making domains. In addition, we show that their representations contain the information necessary to model behavior on the level of individual subjects. Finally, we demonstrate that finetuning on multiple tasks enables large language models to predict human behavior in a previously unseen task. Taken together, these results suggest that large, pre-trained models can be adapted to become generalist cognitive models, thereby opening up new research directions that could transform cognitive psychology and the behavioral sciences as a whole.
\end{abstract}

\section{Introduction}

Large language models are neural networks trained on vast amounts of data to predict the next token for a given text sequence \citep{brown2020language}. These models display many emergent abilities that were not anticipated by extrapolating the performance of smaller models \citep{wei2022emergent}. Their abilities are so impressive and far-reaching that some have argued that they show sparks of general intelligence \citep{bubeck2023sparks}. We may currently witness one of the biggest revolutions in artificial intelligence, but the impact of modern language models is felt far beyond, permeating into education \citep{kasneci2023chatgpt}, medicine \citep{li2023ethics}, and the labor market \citep{eloundou2023gpts}. 

In-context learning -- the ability to extract information from a context and to use that information to improve the production of subsequent outputs -- is one of the defining features of such models. It is through this mechanism that large language models are able to solve a variety of tasks, ranging from translation \citep{brown2020language} to analogical reasoning \citep{webb2022emergent}. Previous work has shown that these models can even successfully navigate when they are placed into classic psychological experiments \citep{binz2023using,coda2023inducing,dasgupta2022language, hagendorff2022machine}. To provide just one example, GPT-3 -- an autoregressive language model designed by OpenAI \citep{brown2020language} -- outperformed human subjects in a sequential decision-making task that required to balance between exploitative and exploratory actions \citep{binz2023using}.

Even though these models show human-like behavioral characteristics in some situations, this is not always the case. In the sequential decision-making task mentioned above, for instance, GPT-3 relied heavily on exploitative strategies, while people applied a combination of elaborate exploration strategies \citep{wilson2014humans}. Moreover, GPT-3 stopped improving after only a few trials, while people continued learning as the task progressed.

\begin{figure}
    \centering
    \includegraphics[scale=0.75]{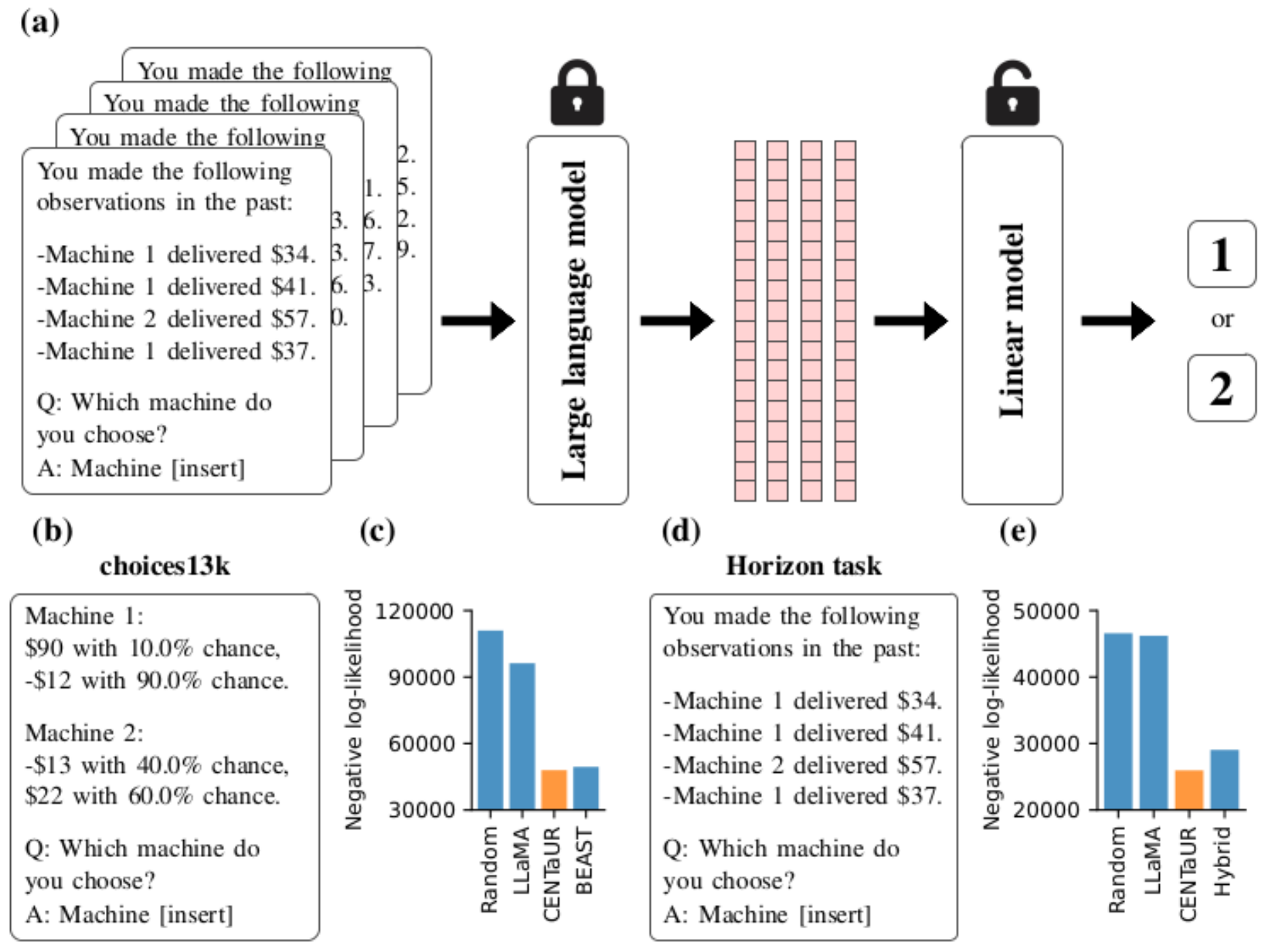}
    \caption{Illustration of our approach and main results. (a) We provided text-based descriptions of psychological experiments to a large language model and extracted the resulting embeddings. We then finetuned a linear layer on top of these embeddings to predict human choices. We refer to the resulting model as CENTaUR. (b) Example prompt for the choices13k data set. (c) Negative log-likelihoods for the choices13k data set. (d) Example prompt for the horizon task. (e) Negative log-likelihoods for the horizon task. Prompts shown in this figure are stylized for readability. Exact prompts can be found in the Supplementary Materials.}
    \label{fig:fig1}
\end{figure}

In the present paper, we investigate whether it is possible to fix the behavioral discrepancy between large language models and humans. To do so, we rely on the idea of finetuning on domain-specific data. This approach has been fruitful across a number of areas \citep{sanh2019distilbert, ouyang2022training} and eventually led to the creation of the term \emph{foundation models} \citep{bommasani2021opportunities} -- models trained on broad data at scale and adapted to a wide range of downstream tasks. In the context of human cognition, such domain-specific data can be readily accessed by tapping the vast amount of behavioral studies that psychologists have conducted over the last century. We made use of this and extracted data sets for several behavioral paradigms which we then used to finetune a large language model. 

We show that this approach can be used to create models that describe human behavior better than traditional cognitive models. We verify this result through extensive model simulations, which confirm that finetuned language models indeed show human-like behavioral characteristics. Furthermore, we find that the embeddings obtained from such models contain the information necessary to capture individual differences. Finally, we highlight that a model finetuned on two tasks is capable of predicting human behavior on a third, hold-out task. Taken together, our work demonstrates that it is possible to turn large language models into cognitive models, thereby opening up completely new opportunities to harvest the power of large language models for building domain-general models of human learning and decision-making.

\section{Finetuned language models beat domain-specific models}

We started our investigations by testing whether it is possible to capture how people make decisions through finetuning a large language model. For our analyses, we relied on the \emph{Large Language Model Meta AI}, or in short: LLaMA \citep{touvron2023llama}. LLaMA is a family of state-of-the-art foundational large language models (with either 7B, 13B, 33B, or 65B parameters) that were trained on trillions of tokens coming from exclusively publicly available data sets. We focused on the largest of these models -- the 65B parameter version -- for the analyses in the main text. LLaMA is publicly available, meaning that researchers are provided with complete access to the network architecture including its pre-trained weights. We utilized this feature to extract embeddings for several cognitive tasks and then finetuned a linear layer on top of these embeddings to predict human choices (see Figure \ref{fig:fig1}a for a visualization). We call the resulting class of models CENTaUR, in analogy to the mythical creature that is half human and half ungulate.

We considered two paradigms that have been extensively studied in the human decision-making literature for our initial analyses: \emph{decisions from descriptions} \citep{kahneman1972subjective} and \emph{decisions from experience} \citep{hertwig2004decisions}. In the former, a decision-maker is asked to choose between one of two hypothetical gambles like the ones shown in Figure \ref{fig:fig1}b. Thus, for both options, there is complete information about outcome probabilities and their respective values. In contrast, the decisions from experience paradigm does not provide such explicit information. Instead, the decision-maker has to learn about outcome probabilities and their respective values from repeated interactions with the task as shown in Figure \ref{fig:fig1}d. Importantly, this modification calls for a change in how an ideal decision-maker should approach such problems: it is not enough to merely exploit currently available knowledge anymore but also crucial to explore options that are unfamiliar \citep{schulz2019algorithmic}. 

For both these paradigms, we created a data set consisting of embeddings and the corresponding human choices. We obtained embeddings by passing prompts that included all the information that people had access to on a given trial through LLaMA and then extracting the hidden activations of the final layer (see Figure \ref{fig:fig1}b and d for example prompts, and the Supplementary Materials for a more detailed description about the prompt generation procedure). We relied on publicly available data from earlier studies in this process. In the decisions from descriptions setting, we used the choices13k data set \citep{peterson2021using}, which is a large-scale data set consisting of over 13,000 choice problems (all in all, 14,711 participants made over one million choices on these problems). In the decisions from experience setting, we used data from the horizon task \citep{wilson2014humans} and a replication study \citep{feng2021dynamics}, which combined include 60 participants making a total of 67,200 choices.

With these two data sets at hand, we fitted a regularized logistic regression model from the extracted embeddings to human choices. In this section, we restricted ourselves to a joint model for all participants, thereby neglecting potential individual differences (but see one of the following sections for an analysis that allows for individual differences). Model performance was measured through the predictive log-likelihood on hold-out data obtained using a 100-fold cross-validation procedure. We standardized all input features and furthermore applied a nested cross-validation for tuning the hyperparameter that controls the regularization strength. Further details are provided in the Materials and Methods section.

We compared the goodness-of-fit of the resulting models against three baselines: a random guessing model, LLaMA without finetuning (obtained by reading out log-probabilities of the pre-trained model), and a domain-specific model (\emph{Best Estimate and Sampling Tools}, or BEAST, for the choices13k data set \citep{erev2017anomalies} and a \emph{hybrid model} \citep{gershman2018deconstructing} that involves a combination of different exploitation and exploration strategies for the horizon task). We found that LLaMA did not capture human behavior well, obtaining a negative log-likelihood (NLL) close to chance-level for the choices13k data set (NLL = $96248.5$) and the horizon task (NLL = $46211.4$). However, finetuning led to models that captured human behavior better than the domain-specific models under consideration. In the choices13k data set, CENTaUR achieved a negative log-likelihood of $48002.3$ while BEAST only achieved a negative log-likelihood of $49448.1$ (see Figure \ref{fig:fig1}c). In the horizon task, CENTaUR achieved a negative log-likelihood of $25968.6$ while the hybrid model only achieved a negative log-likelihood of $29042.5$ (see Figure \ref{fig:fig1}e). Together, these results suggest that the representations extracted from large language models are rich enough to attain state-of-the-art results for modeling human decision-making.

\section{Model simulations reveal human-like behavior}

\begin{figure}
    \centering
    \includegraphics[scale=0.75]{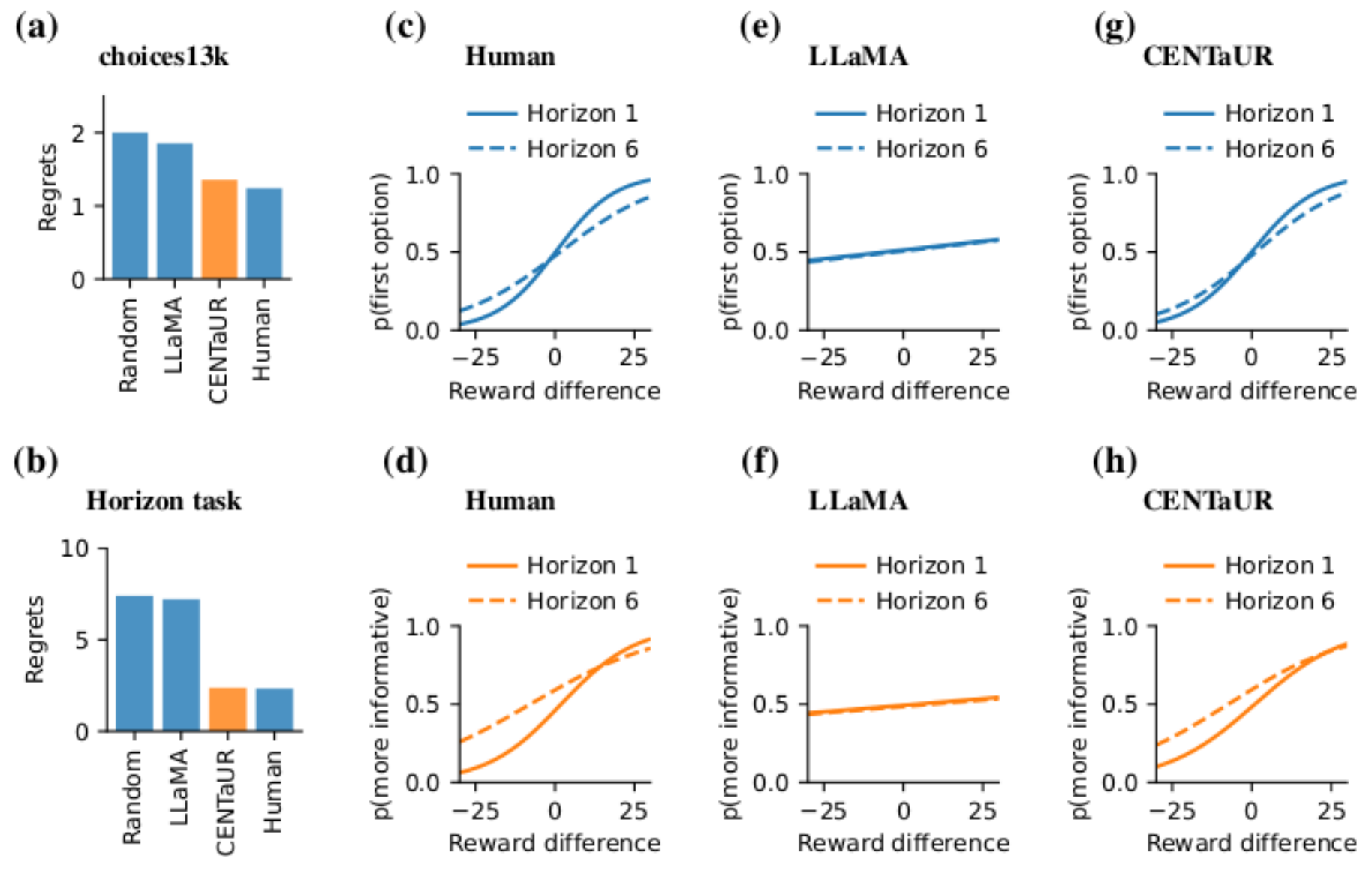}
    \caption{Model simulations. (a) Performance for different models and human participants on the choices13k data set. (b) Performance for different models and human participants on the horizon task. (c) Human choice curves in the equal information condition of the horizon task. (d) Human choice curves in the unequal information condition of the horizon task. (e) LLaMA choice curves in the equal information condition of the horizon task. (f) LLaMA choice curves in the unequal information condition of the horizon task. (g) CENTaUR choice curves in the equal information condition of the horizon task. (h) CENTaUR choice curves in the unequal information condition of the horizon task.}
    \label{fig:fig2}
\end{figure}

We next verified that CENTaUR shows human-like behavioral characteristics. To do so, we simulated the model on the experimental data. Looking at performance, we found that finetuning led to models that closely resemble human performance as shown in Figure \ref{fig:fig2}a and b. For the choices-13k data set, CENTaUR obtained a regret (defined as the difference between the highest possible reward and the reward for the action selected by the model) of $1.35$ (SE $=0.01$), which was much closer to the human regret (M = $1.24$, SE = $0.01$) than the regret of LLaMA (M = $1.85$, SE = $0.01$). The results for the horizon task showed an identical pattern, with CENTaUR (M = $2.38$, SE = $0.01$) matching human regret (M = $2.33$, SE = $0.05$) more closely than LLaMA (M = $7.21$, SE = $0.02$).

In addition to looking at performance, we also inspected choice curves. For this analysis, we took the data from the first free-choice trial in the horizon task and divided it into two conditions: (1) an equal information condition that includes trials where the decision-maker had access to an equal number of observations for both options and (2) an unequal information condition that includes trials where the decision-maker previously observed one option fewer times than the other. We then fitted a separate logistic regression model for each condition with reward difference, horizon, and their interaction as independent variables onto the simulated choices. Earlier studies with human subjects \citep{wilson2014humans} identified the following two main results regarding their exploratory behavior: (1) people’s choices become more random with a longer horizon in the equal information condition (as shown in Figure \ref{fig:fig2}c) and (2) people in the unequal information condition select the more informative option more frequently when the task horizon is longer (as shown in Figure \ref{fig:fig2}d). While LLaMA did not show any of the two effects (see Figure \ref{fig:fig2}e and f), CENTaUR exhibited both of them (see Figure \ref{fig:fig2}g and h), thereby further corroborating that it accurately captures human behavior.

\section{Language model embeddings capture individual differences}

\begin{figure}
    \centering
    \includegraphics[scale=0.75]{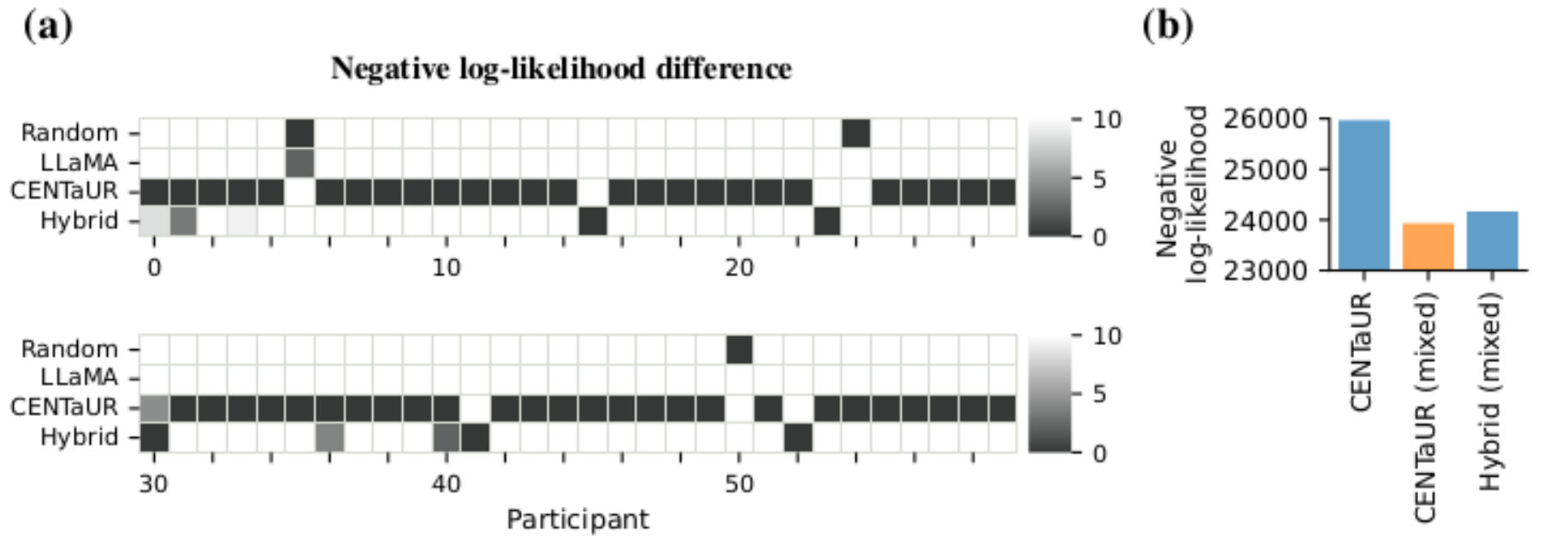}
    \caption{Individual differences. (a) Negative log-likelihood difference to the best-fitting model for each participant. Black highlights the best-fitting model, while white corresponds to a difference larger than ten. (b) Negative log-likelihoods for models that were finetuned using the random-effects structure described in the main text.}
    \label{fig:fig3}
\end{figure}

We also investigated how well CENTaUR describes the behavior of each individual participant. Note that this form of analysis is only possible for the horizon task as choice information on the participant level is not available for the choices13k data set. In total, the majority of participants (N = $52$ out of $60$) was best modeled by CENTaUR (see Figure \ref{fig:fig3}a for a detailed visualization). We furthermore entered the negative log-likelihoods into a random-effects model selection procedure which estimates the probability that a particular model is the most frequent explanation within a set of candidate models \citep{rigoux2014bayesian}. This procedure favored CENTaUR decisively, assigning a probability that it is the most frequent explanation of close to one.

Thus far, we have finetuned LLaMA jointly for all participants. However, people may exhibit individual differences that are not captured by this analysis. To close this gap and test whether LLaMA embeddings can account for individual differences, we incorporated random effects in the finetuned layer. We added a random effect for each participant and embedding dimension while keeping the remaining evaluation procedure the same. Figure \ref{fig:fig3}b illustrates the resulting negative log-likelihoods. Including the random-effect structure improved goodness-of-fit considerably (NLL = $23929.5$) compared to the fixed-effect-only model (NLL = $25968.6$). Furthermore, CENTaUR remained superior to the hybrid model with an identical random-effect structure (NLL = $24166.0$). Taken together, the findings reported in this section highlight that embeddings of large language models contain the information necessary to model behavior on the participant level.

\section{Evaluating goodness-of-fit on hold-out tasks}

\begin{figure}
    \includegraphics[scale=0.75]{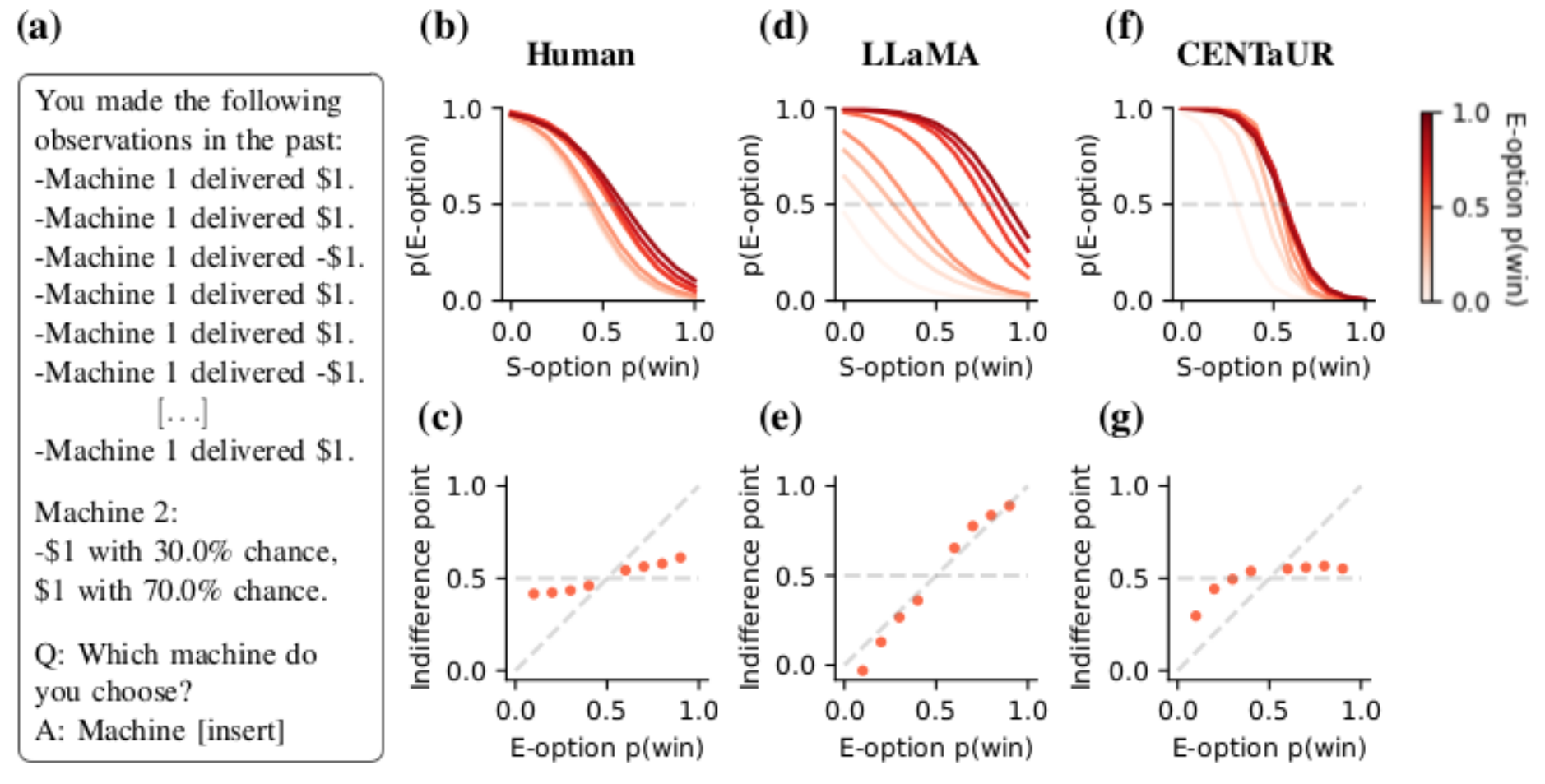}

    \caption{Hold-out task evaluations. (a) Example prompt for the experiential-symbolic task of \citet{garcia2023experiential}. (b) Human choice curves as a function of win probabilities for both options. (c) Human indifference points as a function of win probability for the E-option. Indifferent points express the win probabilities at which a decision-maker is equally likely to select both options. (d) LLaMA choice curves as a function of win probabilities for both options. (e) LLaMA indifference points as a function of win probability for the E-option. (f) CENTaUR choice curves as a function of win probabilities for both options. (g) CENTaUR indifference points as a function of win probability for the E-option.}
    \label{fig:fig4}
\end{figure}

Finally, we examined whether CENTaUR -- after being finetuned on multiple tasks -- is able to predict human behavior in an entirely different task. This evaluation protocol provides a much stronger test for the generalization abilities of our approach. Following our initial analyses, we finetuned a linear layer on top of LLaMA embeddings. However, this time, we fitted a joint model using both the data from the choices13k data set and the horizon task, and then evaluated how well the finetuned model captures human choices on a third task. Further details about the fitting procedure are provided in the Materials and Methods section. For the hold-out task, we considered data from a recent study that provided participants with a choice between one option whose information is provided via a description and another option for which information is provided via a list of experienced outcomes \citep{garcia2023experiential}. Figure \ref{fig:fig4}a shows an example prompt for this experimental paradigm.

Finetuning was generally beneficial for modeling human behavior on the hold-out task: negative log-likelihoods for CENTaUR (NLL = $4521.1$) decreased both in comparison to a random guessing model (NLL = $5977.7$) and LLaMA (NLL = $6307.9$). We were thus curious whether CENTaUR also captures human behavior on a qualitative level. To test this, we took a look at the key insight from the original study: people tend to overvalue options that are provided through a description (symbolic or S-options) over the options that come with a list of experienced outcomes (experiential or E-options) as illustrated in Figure \ref{fig:fig4}b and c. LLaMA does not show this characteristic and instead weighs both option types equally (Figure \ref{fig:fig4}d and e). In contrast to this, CENTaUR shows human-like behavior, taking mostly the S-option into account (Figure \ref{fig:fig4}f and g). This is remarkable because we never presented data from the experiment under consideration during finetuning.

\section{Discussion}

We have demonstrated that large language models can be turned into cognitive models by finetuning their final layer. This process led to models that achieved state-of-the-art performance in two domains. Furthermore, these models were able to capture behavioral differences at the individual participant level. Finally, we have shown that our approach generalizes to previously unseen tasks. In particular, a model that was finetuned on two tasks also exhibited human-like behavior on a third, hold-out task.

These results complement earlier work showing that large language model embeddings allow us to predict behavior and neural activations in linguistic settings \citep{schrimpf2021neural, kumar2022reconstructing, tuckute2023driving, antonello2023scaling}. For example, \citet{schrimpf2021neural} showed that large language models can predict neural and behavioral responses in tasks that involved reading short passages with an accuracy that was close to noise ceiling. While it may be expected that large language models explain human behavior in linguistic domains (after all these models are trained to predict future word occurrences), the observation that these results also transfer to more cognition domains like the ones studied here is highly non-trivial. 

We are particularly excited about one feature of CENTaUR: embeddings extracted for different tasks all lie in a common space. This property allows finetuned large language models to solve multiple tasks in a unified architecture. We have presented preliminary results in this direction, showing that a model finetuned on two tasks can predict human behavior on a third. However, we believe that our current results only hint at the potential of this approach. Ideally, we would like to scale up our approach to finetuning on a larger number of tasks from the psychology literature. If one would include enough tasks in the training set, the resulting system should -- in principle -- generalize to \emph{any} hold-out task. Therefore, our approach provides a path towards a domain-general model of human cognition, which has been the goal of theoreticians for decades \citep{newell1992unified, yang2019task, riveland2022neural, binz2023meta}. We believe that having access to such a model would transform psychology and the behavioral sciences more generally. It could, among other applications, be used to rapidly prototype the outcomes of projected experiments, thereby easing the trial-and-error process of experimental design, or to provide behavioral policy recommendations while avoiding expensive data collection procedures.

Finally, we have to ask ourselves what we can learn about human cognition when finetuning large language models. For now, our insights are limited to the observation that large language model embeddings are rich enough to explain human decision-making. While this is interesting in its own right, it is certainly not the end of the story. Looking beyond the current work, having access to an accurate neural network model of human behavior provides the opportunity to apply a wide range of explainability techniques from the machine learning literature. For instance, we could pick a particular neuron in the embedding and trace back what parts of a particular input sequence excite that neuron using methods such as layer-wise relevance propagation \citep{bach2015pixel, chefer2021transformer}. Thus, our work also opens up a new spectrum of analyses that are not possible when working with human subjects.

To summarize, large language models are an immensely powerful tool for studying human behavior. We believe that our work has only scratched the surface of this potential and there is certainly much more to come.

\paragraph{Acknowledgements:} We like to thank Robert Wilson and Basile Garcia for their help on the horizon task and the experiential-symbolic task respectively, Ido Erev and Eyal Ert for their help with the BEAST model, and Meta AI for making LLaMA accessible to the research community.

\paragraph{Funding:} This work was funded by the Max Planck Society, the Volkswagen Foundation, as well as the Deutsche Forschungsgemeinschaft (DFG,  German  Research  Foundation) under Germany’s  Excellence Strategy–EXC2064/1–390727645.

\paragraph{Data and materials availability:} Data and code for the current study are available  through the GitHub repository \url{https://github.com/marcelbinz/CENTaUR}.

\newpage
\bibliography{bib}
\bibliographystyle{plainnat}

\newpage 

\section*{Supplementary Materials}

\subsection*{Materials and Methods}

\paragraph{Fitting procedure:} 

For the main analyses, we fitted a separate regularized logistic regression model for each data set via a maximum likelihood estimation. Final model performance was measured through the predictive log-likelihood on hold-out data obtained using a 100-fold cross-validation procedure. In each fold, we split the data into a training set ($90\%$), a validation set ($9\%$), and a test set ($1\%$). The validation set was used to identify the parameter $\alpha$ that controls the strength of the $\ell_2$-regularization term. We considered discrete $\alpha$-values of [0, 0.0001, 0.0003, 0.001, 0.003, 0.01, 0.03, 0.1, 0.3, 1.0]. The optimization procedure was implemented in \textsc{PyTorch} \citep{paszke2019pytorch} and used the default LBFGS optimizer \citep{liu1989limited}. For the individual difference analyses, the procedure was identical except that we added a random effect for each participant and embedding dimension.

For the hold-out task analyses, the training set consisted of the concatenated choices13k and horizon task data. To obtain a validation and test set, we split the data of the experiential-symbolic task into eight folds and repeated the previously described fitting procedure for each fold. The validation set was used to identify the parameter $\alpha$ that controls the strength of the $\ell_2$-regularization term and an inverse temperature parameter $\tau^{-1}$. We considered discrete inverse temperature values of [0.05, 0.1, 0.15, 0.2, 0.25, 0.3, 0.35, 0.4, 0.45, 0.5, 0.55, 0.6, 0.65, 0.7, 0.75, 0.8, 0.85, 0.9, 0.95, 1] and $\alpha$-values as described above.

\paragraph{Model simulations:} For the main analyses, we simulated model behavior by sampling from the predictions on the test set. For the hold-out task analyses, we simulated data deterministically based on a median threshold (again using the predictions on the test set). The resulting choice curves were generated by fitting a separate logistic regression model for each possible win probability of the E-option. Each model used the win probability of the S-option and an intercept term as independent variables and the probability of choosing the E-option as the dependent variable.

\paragraph{Baseline models:} For the LLaMA baseline, we fitted an inverse temperature parameter to human choices using the procedure described above. For the BEAST baseline, we relied on the version provided for the choice prediction competition 2018 \citep{plonsky2018and}. We additionally included an error model that selects a random choice with a particular probability. We treated this probability as a free parameter and fitted it using the procedure described above. The hybrid model closely followed the implementation of \citet{gershman2018deconstructing}. We replaced the probit link function with a logit link function to ensure comparability to CENTaUR.

\subsection*{Supplementary Text}

\noindent For the choices13k data set, we prompted each decision independently, thereby ignoring the potential effect of feedback. We used the following template: \\

\begin{adjustwidth}{1cm}{0cm}

\noindent Machine 1 delivers 90 dollars with 10.0\% chance and -12 dollars with 90.0\% chance. \\
Machine 2 delivers -13 dollars with 40.0\% chance and 22 dollars with 60.0\% chance. \\

\noindent Your goal is to maximize the amount of received dollars. \\

\noindent Q: Which machine do you choose? \\
A: Machine [insert] \\

\end{adjustwidth}

\noindent For the horizon task, we prompted each task independently, thereby ignoring potential learning effects across the experiment. We used the following template: \\

\begin{adjustwidth}{1cm}{0cm}
\noindent You made the following observations in the past: \\
 - Machine 1 delivered 34 dollars. \\
 - Machine 1 delivered 41 dollars.  \\
 - Machine 2 delivered 57 dollars. \\
 - Machine 1 delivered 37 dollars. \\

\noindent Your goal is to maximize the sum of received dollars within six additional choices. \\

\noindent Q: Which machine do you choose? \\
A: Machine [insert] \\

\end{adjustwidth}

\noindent For the experiential-symbolic task, we prompted each decision independently and only considered the post-learning phase. We furthermore simplified the observation history by only including the option that is relevant to the current decision. We used the following template: \\

\begin{adjustwidth}{1cm}{0cm}
\noindent You made the following observations in the past: \\
 - Machine 1 delivered 1 dollars. \\
 - Machine 1 delivered 1 dollars. \\
 - Machine 1 delivered -1 dollars. \\
 - Machine 1 delivered 1 dollars. \\
 - Machine 1 delivered 1 dollars. \\
 - Machine 1 delivered -1 dollars. \\
 \hspace*{2.25cm}$\left[\ldots\right]$\\
 - Machine 1 delivered 1 dollars. \\
 
\noindent  Machine 2 delivers -1 dollars with 30.0\% chance and 1 dollars with 70.0\% chance. \\

\noindent  Your goal is to maximize the amount of received dollars. \\

\noindent  Q: Which machine do you choose? \\
A: Machine [insert]

\end{adjustwidth}

\end{document}